\documentclass[letterpaper]{article} % DO NOT CHANGE THIS
\usepackage{aaai2027}  % DO NOT CHANGE THIS
\usepackage[hyphens]{url}  % DO NOT CHANGE THIS
\usepackage{graphicx} % DO NOT CHANGE THIS
\DeclareGraphicsExtensions{.pdf}
\usepackage{multirow}
\usepackage[table]{xcolor}
\definecolor{bestbg}{RGB}{255,220,220}
\definecolor{secondbg}{RGB}{255,235,200}
\newif\iffastcompile
\fastcompilefalse
\iffastcompile
  \setkeys{Gin}{draft=true}
  \ifdefined\pdfcompresslevel
  \fi
  \ifdefined\pdfobjcompresslevel
  \fi
\fi

\usepackage{natbib}  % DO NOT CHANGE THIS AND DO NOT ADD ANY OPTIONS TO IT
\usepackage{caption} % DO NOT CHANGE THIS AND DO NOT ADD ANY OPTIONS TO IT
\usepackage{capt-of}
\usepackage{amsmath}
\usepackage{booktabs}
\DeclareCaptionStyle{ruled}{labelfont=normalfont,labelsep=colon,strut=off} % DO NOT CHANGE THIS

\title{ClearGS: Reliability-Aware Gaussian Splatting from Handheld Videos}
\author {
    Xuanzhi Liu\textsuperscript{\rm 1},
    Xinyi Wu\textsuperscript{\rm 2},
    Hang Pan\textsuperscript{\rm 3},
    Wensi Huang\textsuperscript{\rm 4},\\
    Zhenyao Wu\textsuperscript{\rm 2},
    Ruize Han\textsuperscript{\rm 1},
    Song Wang\textsuperscript{\rm 1}
}

\affiliations{
    \textsuperscript{\rm 1}Shenzhen University of Advanced Technology \quad
    \textsuperscript{\rm 2}HONOR\\
    \textsuperscript{\rm 3}University of New South Wales \quad
    \textsuperscript{\rm 4}Southern University of Science and Technology
}

\begin{document}

\maketitle

\begin{figure*}[t]
\centering
\includegraphics[width=0.98\textwidth]{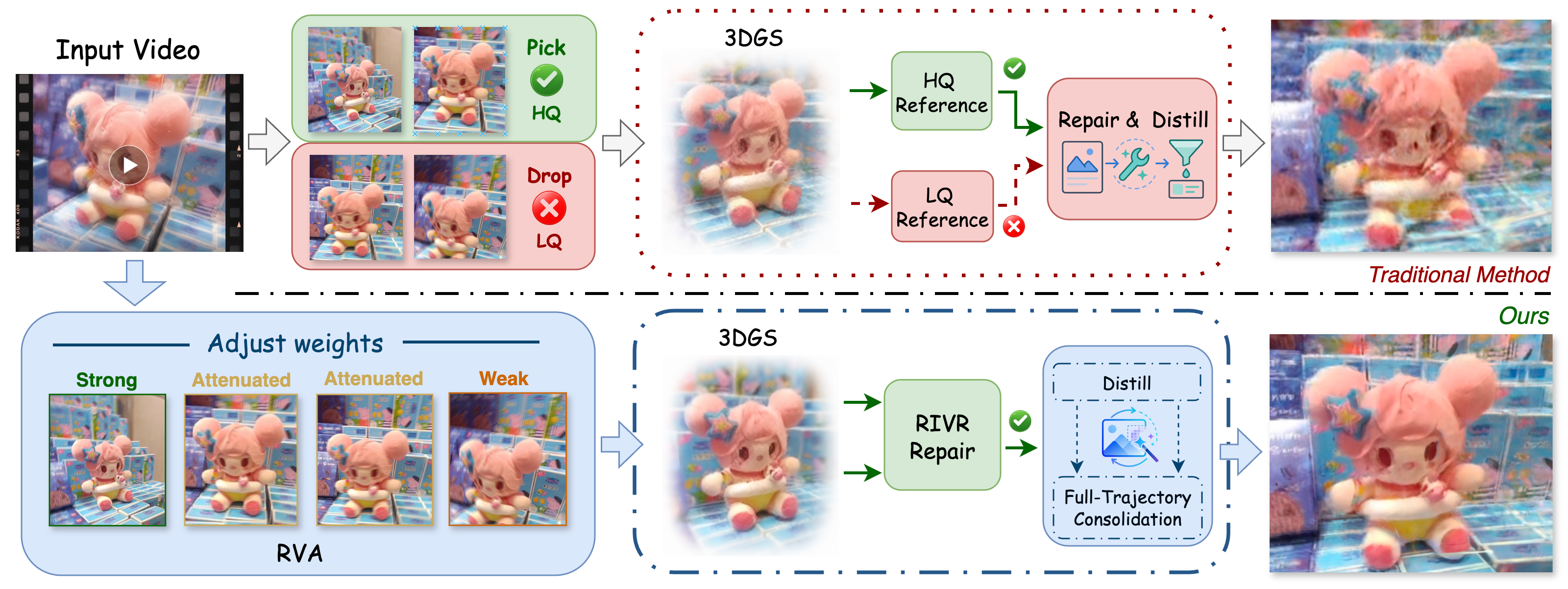}
\caption{Traditional pipelines either train or select handheld-video frames without graded reliability control, then repair degraded views using nearby references or cross-view cues that may share the same artifacts. ClearGS instead uses RVA to separate appearance reliability from geometric utility, and applies no-reference RIVR to add selected repair evidence without matched clean references.}
\label{fig:teaser}
\end{figure*}

\begin{abstract}
We present ClearGS for 3D Gaussian Splatting (3DGS) from handheld videos with uneven viewpoint coverage and mixed frame quality. Rather than selecting frames with binary decisions, ClearGS uses Reliability-aware View Allocation (RVA) to assign graded raw-supervision weights based on appearance reliability, degradation risk, and geometric utility, while weakly reactivating useful suppressed frames to maintain trajectory coverage. Since weighting cannot restore details lost to blur or distortion, ClearGS further introduces Render-Guided In-Video Restoration (RIVR). The current 3DGS render provides a pose-aligned structural candidate, a frozen no-reference restoration expert restores the corresponding raw video observation without any clean reference image, and no-reference perceptual scores select among the render, restored observation, and high-frequency fused candidate. ClearGS then applies Full-Trajectory Repair Consolidation to revisit accepted repairs and preserve details introduced early. On GS2E and GSOTM, ClearGS achieves state-of-the-art overall performance, with consistent CLIP-IQA and MUSIQ gains and LPIPS reductions in most degradation settings, without paired sharp supervision or matched clean references.
\end{abstract}

\section{Introduction}

Recent advances in 3D Gaussian Splatting (3DGS) enable efficient scene reconstruction and high-quality novel-view rendering~\cite{kerbl20233d}, but they assume input views with reasonably consistent appearance and geometry. Handheld videos often break this assumption: frame quality changes along the trajectory, and viewpoint coverage is uneven. Training all frames with equal strength transfers blur, distortion, and inconsistent appearance into the scene. Existing frame- and view-selection methods retain or acquire views according to quality, coverage, uncertainty, or information gain~\cite{polyzos2025activeinitsplat,xue2026uncertainty,chen2026coverage}, but binary keep-or-drop decisions are too coarse for handheld video. A degraded frame may still provide useful geometric evidence; dropping it weakens coverage, while using it at full weight contaminates appearance. We therefore formulate handheld-video 3DGS as a mixed-quality supervision problem that separates appearance reliability from geometric utility.

ClearGS first addresses this problem with Reliability-aware View Allocation (RVA). For each frame, RVA estimates appearance reliability, degradation risk, and geometric utility, then assigns an ordered raw-supervision weight. Reliable low-risk frames receive strong supervision, geometrically useful observations receive attenuated supervision, and unreliable low-utility frames are suppressed. To avoid coverage holes, RVA also weakly reactivates useful suppressed frames in under-covered trajectory regions, preserving viewpoint coverage without letting degraded appearance dominate the initial reconstruction.

Weighted supervision alone cannot recover details already lost to blur or distortion. Existing restoration--distillation methods often depend on nearby RGB references, aggregated sparse-view features, or depth-guided correspondences~\cite{wu2025difix3d+,liu20243dgs,wu2025genfusion,yin2025gsfixer,cao2026geoquery}. These cues are unreliable in handheld videos: neighboring frames may suffer from the same degradation, and corrupted appearance can be propagated across views. Render-Guided In-Video Restoration (RIVR) extends ClearGS from raw supervision to repaired evidence. For each target pose, RIVR renders the current scene as a pose-aligned structural candidate and restores the corresponding raw video observation with a frozen no-reference restoration expert, without using any clean reference image or paired sharp target. RIVR also constructs a detail-fused candidate by combining the rendered base with restored high-frequency details. It then selects among the render, restored observation, and fused candidate using no-reference perceptual scores, and appends the selected result as a pseudo observation for later distillation.

Progressive distillation can weaken details introduced by repairs accepted early in the trajectory. ClearGS therefore performs Full-Trajectory Repair Consolidation, which reuses the latest accepted repair at each target pose without sampling new poses or generating new candidates. Together, RVA, RIVR, and Full-Trajectory Repair Consolidation preserve reliable captured evidence, recover video-grounded details, and maintain weakly observed regions. Experiments on GS2E~\cite{li2025gs2e} and GSOTM~\cite{seiskari2024gaussian} show state-of-the-art overall performance, with sharper boundaries and more stable textures under mixed-quality observations.

Our contributions are threefold:
\begin{itemize}
    \item We formulate 3DGS reconstruction from handheld videos as a mixed-quality supervision problem that separates appearance reliability, degradation risk, and geometric utility.
    \item We introduce RVA, which assigns graded raw-supervision weights and preserves trajectory coverage through weak reactivation without allowing degraded observations to dominate scene appearance.
    \item We introduce RIVR, a no-reference in-video restoration and distillation strategy that selects repair evidence without clean references and consolidates accepted repairs through Full-Trajectory Repair Consolidation.
\end{itemize}

\begin{figure*}[!t]
\centering
\includegraphics[width=0.95\textwidth]{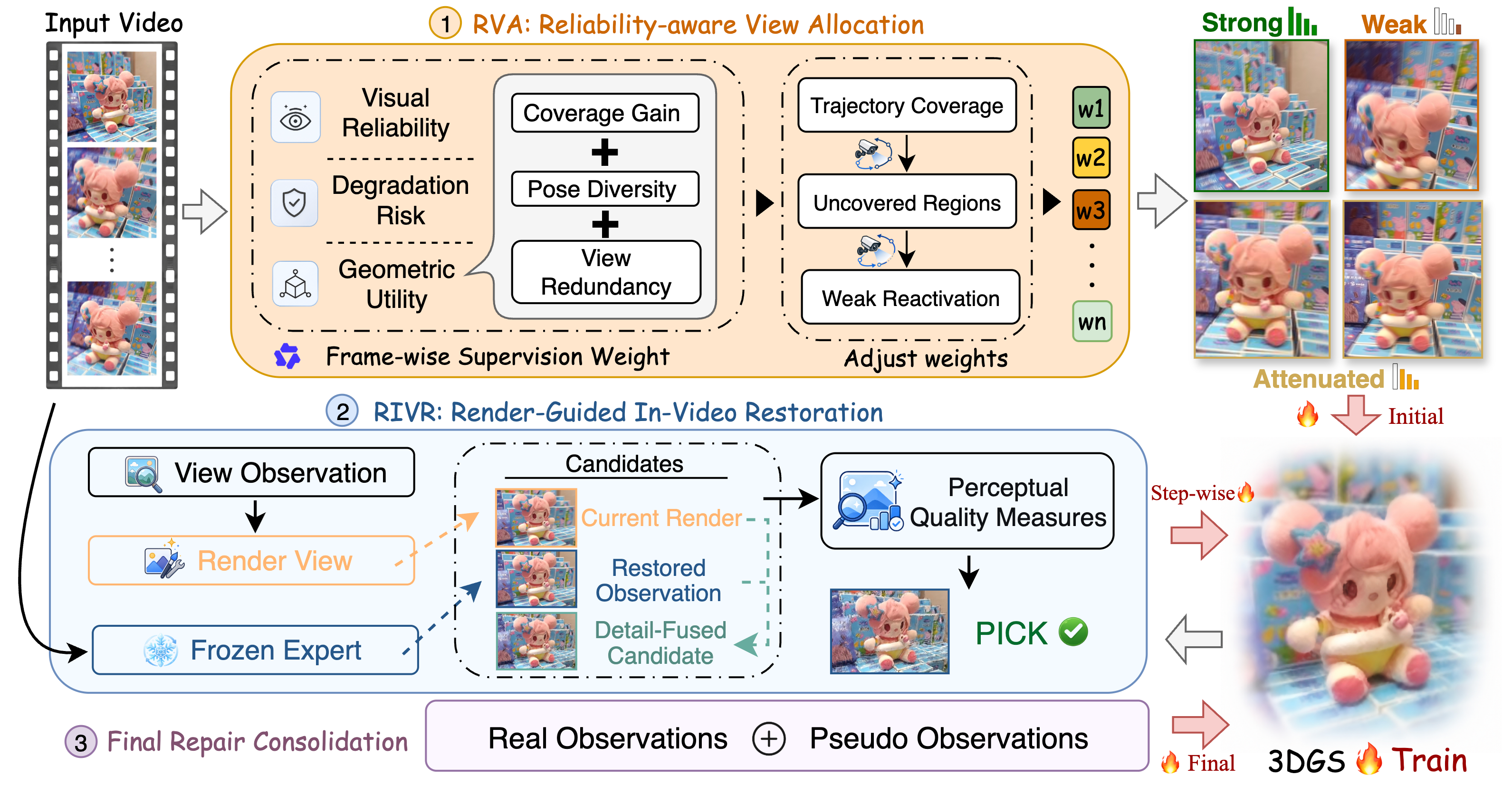}
\caption{Overview of ClearGS. RVA assigns frame-wise raw-supervision weights from appearance reliability, degradation risk, and geometric utility, and weakly reactivates useful suppressed frames in uncovered trajectory regions. RIVR selects among the current render, restored raw observation, and detail-fused candidate using no-reference quality scores, then distills accepted pseudo observations with real observations.}
\label{fig:method_overview}
\end{figure*}

\section{Related Work}
\subsection{Scene Reconstruction from Blurry Images and Handheld Videos}

Blur-aware radiance-field methods recover sharp scenes by learning spatially varying kernels, imposing physical priors, or estimating exposure-time camera trajectories~\cite{ma2022deblur,lee2023dp,wang2023bad,lee2023exblurf}. Recent 3DGS methods model blur through Gaussian covariance manipulation, per-pixel kernels, or continuous camera motion~\cite{lee2024deblurring,zhao2024bad,peng2024bags,lee2025comogaussian}, and handheld-video methods further compensate for rolling shutter and pose errors~\cite{seiskari2024gaussian,zhang2026unblur}. These methods explain degradation through image-formation or camera-motion models. ClearGS addresses a different failure mode: frame quality and viewpoint coverage vary along a fixed handheld trajectory. It controls the appearance influence of each frame while retaining observations that still provide useful geometric evidence.

\subsection{Frame and View Selection for Reconstruction}

Frame and active view selection methods retain informative observations or acquire new viewpoints according to image quality, coverage, uncertainty, or information gain~\cite{polyzos2025activeinitsplat,xue2026uncertainty,chen2026coverage}. Their goal is often to improve acquisition efficiency or expose under-observed geometry. In fixed handheld videos, however, binary selection cannot use the geometric value of a degraded frame without also accepting its unreliable appearance. RVA separates appearance reliability, degradation risk, and geometric utility, assigning graded raw-supervision weights rather than hard decisions. It also weakly reactivates useful suppressed frames to avoid trajectory coverage holes.

\subsection{3DGS Recovery from Imperfect Observations}

Methods for sparse or imperfect observations stabilize 3DGS using depth priors, multi-view constraints, structural regularization, and Gaussian control~\cite{chung2024depth,xiong2023sparsegs,zhu2024fsgs,li2024dngaussian,xu2024mvpgs,zhang2024cor,park2025dropgaussian,chen2026quantifying,song2025d}. Generative methods repair rendered views or add pseudo-observations~\cite{wu2024reconfusion,liu20243dgs,wu2025genfusion,shen2026sparse,jiang2026freescale,li2026syncfix,ni2025g4splat,zhu2026gaussfusion}, while recent 3DGS repair methods often condition restoration on nearby RGB references, sparse-view features, or depth-guided correspondences~\cite{wu2025difix3d+,yin2025gsfixer,cao2026geoquery}. These cues can be unreliable in mixed-quality handheld videos, where neighboring frames may share the same degradation and corrupted appearance can propagate across views. RIVR instead restores the corresponding raw video observation with a frozen no-reference restoration expert, using no clean reference image or paired sharp target. The current scene render provides a pose-aligned structural candidate for selection and fusion, and accepted repairs are later reused through Full-Trajectory Repair Consolidation.

\section{Method}

\subsection{Problem Definition}
We consider a handheld posed video sequence
\begin{equation}
\mathcal{V}
=
\{(\mathbf{I}_t,\boldsymbol{\pi}_t)\}_{t=1}^{T},
\label{eq:task_input}
\end{equation}
where $\mathbf{I}_t$ is an input video frame and
$\boldsymbol{\pi}_t$ denotes its camera parameters. Such videos contain
observations of heterogeneous quality: clear frames coexist with frames
affected by motion blur, defocus, distortion, or other degradations.
Moreover, viewpoint coverage is often uneven along the camera trajectory. A
degraded frame may therefore be unsafe as appearance supervision while still
providing the only observation of a weakly covered region.

Our goal is to reconstruct a 3DGS scene $\mathcal{G}^{\star}$ from
$\mathcal{V}$ that faithfully renders both observed and novel views. During
reconstruction, only the input video is available; paired sharp targets and
high-quality reference images matched to the target pose are unavailable.
The central challenge is to preserve the geometric coverage offered by
degraded observations without directly writing their unreliable appearance
into the scene representation.

\begin{table*}[t]
\centering
\scriptsize
\caption{RVA notation, parameter settings, and supervision policy. The standard supervision conditions are evaluated from top to bottom. Afterward, trajectory coverage compensation may reactivate a suppressed frame in an uncovered trajectory bin with a fixed weak-supervision weight.}
\label{tab:rva_params}
\setlength{\tabcolsep}{3.0pt}
\renewcommand{\arraystretch}{1.02}
\resizebox{\textwidth}{!}{%
\begin{tabular}{@{}llll@{}}
\toprule
Part & Quantity / role & Definition or condition & Parameters / weight \\
\midrule
Appearance & $\mathbf{r}_i$ & ordered quality cues: sharpness, structural clarity, texture reliability, artifact cleanliness, exposure quality & $\mathbf{w}_R=(0.30,0.25,0.20,0.15,0.10)$ \\
& $\mathbf{b}_i$ & ordered degradation cues: motion blur, defocus, ghosting, exposure failure, unreliable texture, artifacts & $\mathbf{w}_B=(0.25,0.15,0.20,0.15,0.15,0.10)$ \\
& $R_i$ & appearance reliability score & $R_i=\mathbf{w}_R^\top\mathbf{r}_i$ \\
& $B_i$ & degradation risk score & $B_i=\mathbf{w}_B^\top\mathbf{b}_i$ \\
& $\mathcal{A}$ & appearance-reliable frame set & $\tau_R=0.70,\ \tau_B=0.60$ \\
\midrule
Geometry & $d_{ij}^{c}$ & normalized camera-center distance between frames $i$ and $j$ & Euclidean distance normalized to $[0,1]$ \\
& $d_{ij}^{v}$ & normalized viewing-direction difference between frames $i$ and $j$ & viewing-direction angle normalized to $[0,1]$ \\
& $D_i$ & $\displaystyle D_i=\max_{j\in\mathcal{A}}\left(\gamma_c d_{ij}^{c}+\gamma_v d_{ij}^{v}\right)$ & $(\gamma_c,\gamma_v)=(0.65,0.35)$ \\
& $O_i$ & $\displaystyle O_i=\max_{j\in\mathcal{A}}\frac{\left|\mathcal{P}_i\cap\mathcal{P}_j\right|}{\left|\mathcal{P}_i\cup\mathcal{P}_j\right|}$ & -- \\
& $G_i$ & geometric utility score & $(\alpha,\beta,\eta)=(0.45,0.35,0.30)$ \\
\midrule
Trajectory compensation & Selection coefficients & coefficients in Eq.~\eqref{eq:tca_selection} & $(\lambda_G,\lambda_R,\lambda_B,\lambda_Z)=(0.35,0.25,0.30,0.10)$ \\
& Reactivated supervision & suppressed frame selected as $i_m^\ast$ in an uncovered trajectory bin & $w_{i_m^\ast}=0.12$ \\
\midrule
Supervision assignment & Strong & $R_i\geq0.70,\ B_i\leq0.60$ & $w_i=0.80+0.20R_i$ \\
& Geometrically attenuated & $G_i\geq0.60,\ B_i\leq0.80$ & $w_i=0.30+0.30(0.35R_i+0.65G_i)$ \\
& Auxiliary attenuated & $R_i\geq0.35,\ G_i\geq0.25,\ B_i\leq0.75$ & $w_i=0.30+0.30(0.50R_i+0.50G_i)$ \\
& Coverage-reactivated & initially suppressed but selected as $i_m^\ast$ & $w_i=0.12$ \\
& Suppressed & otherwise and not selected for coverage compensation & $w_i=0$ \\
\bottomrule
\end{tabular}%
}
\end{table*}

\subsection{Overview}

ClearGS follows a reliability-first principle for handheld video reconstruction: we first establish a reliable initial scene before introducing repaired observations. RVA evaluates frames from appearance and geometric perspectives, allocates reliability-aware supervision, and preserves necessary trajectory coverage to initialize $\mathcal{G}_{0}$. Starting from $\mathcal{G}_{0}$, RIVR forms three target-pose candidates: the current 3DGS render, the restored raw video frame, and a fused result that combines the rendered base with restored high-frequency details. The selected candidate is appended as an additional pseudo observation and optimized together with the original video supervision, complementing rather than replacing the captured frames. Finally, full-trajectory repair consolidation gathers the accepted pseudo observations across the trajectory and performs a final joint distillation pass, producing the optimized scene $\mathcal{G}^{\star}$.

\subsection{Reliability-aware View Allocation (RVA)}
Handheld video reconstruction is influenced by both the appearance reliability of individual frames and their geometric utility for scene reconstruction. Frames with reliable appearance provide stronger photometric supervision, while frames captured from complementary viewpoints provide additional geometric constraints. Therefore, ClearGS evaluates each frame from both appearance and geometric perspectives and assigns reliability-aware supervision weights, instead of simply retaining or discarding frames based on a single criterion
~\cite{polyzos2025activeinitsplat,xue2026uncertainty,chen2026coverage}.

\subsubsection{Appearance Evidence}
We first evaluate whether a frame provides reliable appearance information. For each input frame $\mathbf{I}_i$, RVA uses a frozen Qwen3-VL~\cite{bai2025qwen3} as an image-based evaluator. Qwen3-VL predicts appearance quality cues and degradation patterns, which are aggregated into two complementary scores. Quality-related cues, including sharpness, structural clarity, texture reliability, artifact cleanliness, and exposure quality, are combined into a reliability score $R_i$, indicating
whether the frame can provide trustworthy appearance supervision. Degradation patterns, including motion blur, defocus, ghosting, exposure failure, unreliable texture, and artifacts, are combined into a degradation risk score $B_i$, indicating the likelihood of introducing unreliable appearance into the reconstruction. Frames with high $R_i$ and low $B_i$ are considered appearance-reliable frames $\mathcal{A}$:
\begin{equation}
\mathcal{A}=\{j\mid R_j\geq\tau_R,\ B_j\leq\tau_B\},
\end{equation}
where $\tau_R$ and $\tau_B$ are
predefined thresholds for reliability and degradation risk, respectively. The score aggregation weights and threshold values are summarized in Table~\ref{tab:rva_params}.

\subsubsection{Geometric Evidence}
Besides appearance information, ClearGS considers the geometric contribution of each frame to scene reconstruction. A frame with degraded appearance may still provide valuable observations from complementary viewpoints. We estimate geometric utility based on the COLMAP sparse reconstruction and camera poses.

Given the reliable frame set $\mathcal{A}$ obtained from appearance assessment, we construct a reference coverage by merging their visible 3D points:
\begin{equation}
\mathcal{P}_{\mathrm{rel}}
=
\bigcup_{j\in\mathcal{A}}\mathcal{P}_j,
\end{equation}
where $\mathcal{P}_j$ denotes the set of COLMAP 3D points observed by frame $j$. This coverage represents scene regions supported by reliable appearance observations.

For each frame $i$, we measure geometric utility using three complementary cues: coverage gain $C_i$, pose diversity $D_i$, and visibility redundancy $O_i$. Coverage gain measures additional 3D points observed by frame $i$ beyond $\mathcal{P}_{\mathrm{rel}}$:
\begin{equation}
C_i=
\frac{
\sum_{p\in\mathcal{P}_i\setminus\mathcal{P}_{\mathrm{rel}}}
\frac{1}{\sqrt{n_p}}
}
{
\sum_{p\in\mathcal{P}_i}
\frac{1}{\sqrt{n_p}}
},
\end{equation}
where $n_p$ denotes the number of COLMAP images observing point $p$, assigning larger weights to sparsely observed points. Pose diversity measures the complementarity of the camera pose of frame $i$ with reliable observations, while visibility redundancy measures its overlap with existing coverage.
The geometric utility is computed as:
\begin{equation}
G_i=
\operatorname{clip}_{[0,1]}
\left(
\alpha C_i+\beta D_i-\eta O_i
\right),
\label{eq:vista_geometry}
\end{equation}
where $\alpha$, $\beta$, and $\eta$ balance the three geometric cues. A higher $G_i$ indicates that the frame provides complementary geometric evidence and receives higher priority during supervision allocation. The formulations of $D_i$ and $O_i$, together with their corresponding weights, are summarized in Table~\ref{tab:rva_params}.

\subsubsection{Trajectory Coverage Compensation}
However, independent frame-wise decisions may suppress neighboring frames in the same trajectory region, causing local coverage gaps. To preserve sufficient camera coverage during scene initialization, RVA further performs trajectory coverage compensation. Specifically, the trajectory is partitioned into pose-aware bins according to camera motion. If a bin already contains active frames, it remains unchanged. Otherwise, RVA reactivates one suppressed frame $i_m^\ast$:
\begin{equation}
i_m^\ast=
\arg\max_{i\in\mathcal{S}_m}
\left(
\lambda_GG_i+
\lambda_RR_i+
\lambda_B(1-B_i)+
\lambda_ZZ_i
\right),
\label{eq:tca_selection}
\end{equation}
where $\mathcal{S}_m$ denotes the suppressed frames in the uncovered trajectory bin. $Z_i$ denotes the trajectory centrality score, where larger values favor frames closer to the bin center. The trajectory-compensation coefficients and weak reactivation weight are summarized in Table~\ref{tab:rva_params}.

Finally, RVA assigns each frame a supervision role according to
Table~\ref{tab:rva_params}. The resulting supervision set, including both allocated frames and trajectory-compensated frames, is used to initialize $\mathcal{G}_0$.

\subsection{Render-Guided In-Video Restoration (RIVR)}

After RVA builds the initial scene $\mathcal{G}_0$, some details from degraded frames may still be missing or weak in the reconstruction. RIVR addresses this problem with no-reference in-video restoration. It does not require a matched clean image, a paired sharp target, nearby RGB references, aggregated cross-view features, or explicit geometric correspondences~\cite{wu2025difix3d+,liu20243dgs,wu2025genfusion,yin2025gsfixer,cao2026geoquery}. Instead, it uses two signals available at the same target pose: the current 3DGS render and the corresponding raw video observation.

For each target frame $t$, RIVR renders the current scene as $\mathbf{R}_t$ and retrieves the raw observation $\mathbf{O}_t$. The raw observation, rather than the rendered image, is processed by a frozen Restormer motion-deblurring expert~\cite{zamir2022restormer}, producing $\mathbf{D}_t$. The render provides pose-aligned scene structure, but it is not treated as a clean reference. The restored observation provides target-view appearance evidence recovered from the original video frame. RIVR then forms the candidate set summarized in Table~\ref{tab:rivr_candidates}.

\begin{table}[t]
\centering
\scriptsize
\caption{RIVR candidates for no-reference in-video restoration.}
\label{tab:rivr_candidates}
\resizebox{\linewidth}{!}{%
\begin{tabular}{lll}
\toprule
Candidate & Construction & Role \\
\midrule
Render & $\mathbf{R}_t$ & Pose-aligned structural candidate \\
Restored & $\mathbf{D}_t$ & Restoration of the raw target observation \\
Fused & $\mathbf{F}_t=\mathbf{R}_t+\mathbf{D}_t-\mathcal{B}_{\sigma}(\mathbf{D}_t)$ & Render base with restored high-frequency detail \\
\bottomrule
\end{tabular}%
}
\end{table}

Because clean references are unavailable during reconstruction, RIVR selects among the candidates with a reference-free rule. MUSIQ~\cite{ke2021musiq} is used as the primary score because it responds to blur, detail loss, and unnatural local structure, while CLIP-IQA is used only when MUSIQ scores are close. The selected candidate is appended as a pseudo observation for later distillation. Thus, restoration evidence is used only after being chosen from render, restored, and fused candidates, without assuming any clean target image or cross-view reference.

\subsection{Joint Distillation}

Following restoration-guided reconstruction methods~\cite{wu2025difix3d+,yin2025gsfixer,cao2026geoquery}, ClearGS distills accepted RIVR outputs with the active real views selected by RVA. At round $r$, the training set is
\begin{equation}
\mathcal{D}^{(r)}
=
\mathcal{D}_{\mathrm{real}}
\cup
\bigcup_{\ell=1}^{r}
\mathcal{D}_{\mathrm{pseudo}}^{(\ell)}.
\label{eq:joint_view_set}
\end{equation}
Repaired views are appended rather than used to replace real observations, keeping captured video evidence while adding selected restoration details.

Since later rounds may weaken repairs accepted early, ClearGS applies Full-Trajectory Repair Consolidation after all frames are processed. It reuses the accepted repairs from the whole trajectory, keeps only the latest repair for repeated target frames, and introduces no new poses or repair candidates.

\section{Experiment}
\subsection{Experimental Setup}

We evaluate ClearGS on GS2E~\cite{li2025gs2e} and Gaussian Splatting on the Move (GSOTM)~\cite{seiskari2024gaussian}.
GS2E contains ten scenes at each of three synthetic blur levels: slight,
medium, and severe. GSOTM complements this protocol with four camera-motion
degradation settings: motion blur (MB), rolling shutter (RS), their combination
(MB+RS), and pose noise (PN). Every method receives the same degraded posed
observations. Neither sharp ground-truth images nor matched high-quality
reference images are available during view allocation, reconstruction, repair
selection, or distillation; sharp targets are used only for final
reference-based evaluation.
\begin{table*}[!t]
\centering
\caption{Perceptual comparison on GS2E and GSOTM at observed input poses. GS2E averages ten scenes per blur level, and GSOTM averages four scenes per degradation type. LPIPS uses sharp targets; CLIP-IQA and MUSIQ are no-reference metrics. Best and second-best results are highlighted.}
\label{tab:main_comparison}
\scriptsize
\setlength{\tabcolsep}{2.0pt}
\renewcommand{\arraystretch}{1.08}
\resizebox{\textwidth}{!}{%
\begin{tabular}{@{}l|ccccccc|ccccccc|ccccccc@{}}
\toprule
\multirow{3}{*}{Method}
& \multicolumn{7}{c|}{LPIPS$\downarrow$}
& \multicolumn{7}{c|}{CLIP-IQA$\uparrow$}
& \multicolumn{7}{c}{MUSIQ$\uparrow$} \\
\cmidrule(lr){2-8}
\cmidrule(lr){9-15}
\cmidrule(l){16-22}

& \multicolumn{3}{c|}{GS2E}
& \multicolumn{4}{c|}{GSOTM}
& \multicolumn{3}{c|}{GS2E}
& \multicolumn{4}{c|}{GSOTM}
& \multicolumn{3}{c|}{GS2E}
& \multicolumn{4}{c}{GSOTM} \\
\cmidrule(lr){2-4}
\cmidrule(lr){5-8}
\cmidrule(lr){9-11}
\cmidrule(lr){12-15}
\cmidrule(lr){16-18}
\cmidrule(l){19-22}

& Slight & Medium & Severe
& MB & RS & MB+RS & PN
& Slight & Medium & Severe
& MB & RS & MB+RS & PN
& Slight & Medium & Severe
& MB & RS & MB+RS & PN \\
\midrule

All-view $G_0$
& 0.397
& 0.414
& 0.430
& 0.323
& 0.334
& 0.419
& 0.476
& \cellcolor{secondbg}0.266
& \cellcolor{secondbg}0.261
& 0.256
& 0.194
& 0.212
& 0.196
& 0.227
& 34.82
& 32.02
& 30.11
& 36.75
& 50.25
& 34.51
& 34.06 \\

Select-G $G_0$ (Ours)
& 0.386
& 0.401
& 0.415
& 0.285
& 0.333
& 0.393
& 0.472
& \cellcolor{secondbg}0.266
& 0.256
& \cellcolor{secondbg}0.261
& 0.201
& 0.211
& 0.187
& 0.224
& 36.53
& 34.51
& 34.22
& \cellcolor{secondbg}42.31
& 50.55
& 39.46
& 35.42 \\
\midrule

DiFiX3D~\cite{wu2025difix3d+}
& \cellcolor{secondbg}0.349
& \cellcolor{secondbg}0.372
& \cellcolor{secondbg}0.388
& \cellcolor{secondbg}0.224
& \cellcolor{bestbg}\textbf{0.224}
& \cellcolor{secondbg}0.322
& \cellcolor{secondbg}0.249
& 0.258
& 0.255
& 0.253
& \cellcolor{secondbg}0.231
& \cellcolor{secondbg}0.253
& 0.213
& 0.230
& \cellcolor{secondbg}41.53
& \cellcolor{secondbg}39.03
& 36.93
& 41.64
& \cellcolor{secondbg}59.77
& 39.94
& \cellcolor{secondbg}51.64 \\

GeoQuery~\cite{cao2026geoquery}
& 0.350
& 0.373
& 0.389
& 0.231
& \cellcolor{bestbg}\textbf{0.224}
& 0.332
& \cellcolor{secondbg}0.249
& 0.256
& 0.253
& 0.252
& 0.230
& 0.250
& \cellcolor{secondbg}0.216
& \cellcolor{secondbg}0.231
& 41.33
& 38.70
& \cellcolor{secondbg}37.20
& 41.86
& 59.50
& \cellcolor{secondbg}40.02
& 51.27 \\

SyncFix~\cite{li2026syncfix}
& 0.353
& 0.377
& 0.395
& 0.253
& 0.227
& 0.334
& 0.250
& 0.253
& 0.246
& 0.242
& 0.225
& 0.249
& 0.208
& 0.227
& 39.99
& 36.76
& 34.44
& 37.90
& 59.00
& 35.67
& 50.26 \\
\midrule

ClearGS (Ours)
& \cellcolor{bestbg}\textbf{0.333}
& \cellcolor{bestbg}\textbf{0.342}
& \cellcolor{bestbg}\textbf{0.350}
& \cellcolor{bestbg}\textbf{0.193}
& \cellcolor{secondbg}0.225
& \cellcolor{bestbg}\textbf{0.302}
& \cellcolor{bestbg}\textbf{0.197}
& \cellcolor{bestbg}\textbf{0.288}
& \cellcolor{bestbg}\textbf{0.287}
& \cellcolor{bestbg}\textbf{0.293}
& \cellcolor{bestbg}\textbf{0.314}
& \cellcolor{bestbg}\textbf{0.296}
& \cellcolor{bestbg}\textbf{0.243}
& \cellcolor{bestbg}\textbf{0.267}
& \cellcolor{bestbg}\textbf{44.67}
& \cellcolor{bestbg}\textbf{43.57}
& \cellcolor{bestbg}\textbf{43.18}
& \cellcolor{bestbg}\textbf{62.03}
& \cellcolor{bestbg}\textbf{64.25}
& \cellcolor{bestbg}\textbf{52.64}
& \cellcolor{bestbg}\textbf{59.07} \\

\bottomrule
\end{tabular}%
}
\end{table*}

\begin{table*}[!t]
\centering
\caption{Ablation of RVA coverage allocation. S/M/V are GS2E blur levels; MB/RS/MB+RS/PN are GSOTM degradation settings.}
\label{tab:rva_coverage_ablation}
\scriptsize
\setlength{\tabcolsep}{2.0pt}
\renewcommand{\arraystretch}{1.05}
\resizebox{\textwidth}{!}{%
\begin{tabular}{@{}lccccccccccccccccccccc@{}}
\toprule
\multirow{3}{*}{Variant}
& \multicolumn{7}{c}{PSNR$\uparrow$}
& \multicolumn{7}{c}{SSIM$\uparrow$}
& \multicolumn{7}{c}{LPIPS$\downarrow$} \\
\cmidrule(lr){2-8}
\cmidrule(lr){9-15}
\cmidrule(l){16-22}
& \multicolumn{3}{c}{GS2E}
& \multicolumn{4}{c}{GSOTM}
& \multicolumn{3}{c}{GS2E}
& \multicolumn{4}{c}{GSOTM}
& \multicolumn{3}{c}{GS2E}
& \multicolumn{4}{c}{GSOTM} \\
\cmidrule(lr){2-4}
\cmidrule(lr){5-8}
\cmidrule(lr){9-11}
\cmidrule(lr){12-15}
\cmidrule(lr){16-18}
\cmidrule(l){19-22}
& S & M & V
& MB & RS & MB+RS & PN
& S & M & V
& MB & RS & MB+RS & PN
& S & M & V
& MB & RS & MB+RS & PN \\
\midrule
RVA w/o cov.
& \cellcolor{bestbg}\textbf{20.123} & 19.538 & 19.499
& 21.621 & 18.010 & 17.624 & 18.413
& \cellcolor{bestbg}\textbf{0.607} & 0.592 & 0.589
& 0.730 & \cellcolor{bestbg}\textbf{0.530} & 0.534 & \cellcolor{bestbg}\textbf{0.573}
& 0.388 & 0.413 & \cellcolor{bestbg}\textbf{0.415}
& 0.290 & 0.339 & \cellcolor{bestbg}\textbf{0.387} & \cellcolor{bestbg}\textbf{0.472} \\
Full RVA
& 19.934 & \cellcolor{bestbg}\textbf{20.076} & \cellcolor{bestbg}\textbf{20.011}
& \cellcolor{bestbg}\textbf{22.545} & \cellcolor{bestbg}\textbf{18.124} & \cellcolor{bestbg}\textbf{18.010} & \cellcolor{bestbg}\textbf{18.458}
& 0.606 & \cellcolor{bestbg}\textbf{0.609} & \cellcolor{bestbg}\textbf{0.608}
& \cellcolor{bestbg}\textbf{0.739} & \cellcolor{bestbg}\textbf{0.530} & \cellcolor{bestbg}\textbf{0.541} & \cellcolor{bestbg}\textbf{0.573}
& \cellcolor{bestbg}\textbf{0.386} & \cellcolor{bestbg}\textbf{0.401} & \cellcolor{bestbg}\textbf{0.415}
& \cellcolor{bestbg}\textbf{0.285} & \cellcolor{bestbg}\textbf{0.333} & 0.393 & \cellcolor{bestbg}\textbf{0.472} \\
\bottomrule
\end{tabular}%
}
\end{table*}

\subsection{Effect of Reliability-aware View Allocation}

\begin{figure}[!t]
\centering
\includegraphics[width=0.96\linewidth]{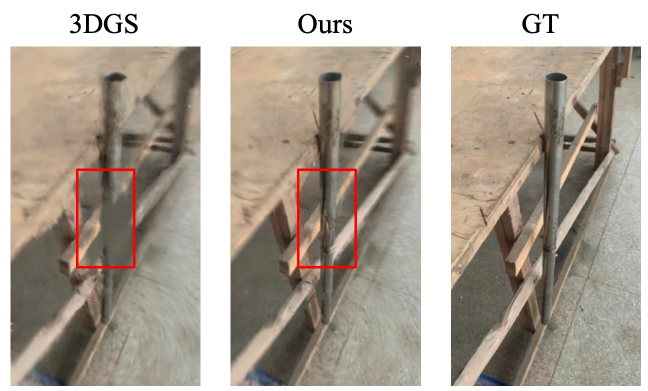}
\caption{Qualitative comparison of renderings at observed input camera poses produced by
all-view 3DGS and the Select-view initialization of ClearGS. Red boxes highlight regions
where our reconstruction better preserves thin structures, object boundaries,
and local appearance relative to all-view supervision.}
\label{fig:view_allocation_qualitative}
\end{figure}

We isolate the reconstruction stage by comparing all-view $\mathcal{G}_0$ with
RVA-selected $\mathcal{G}_0$. As shown in
Figure~\ref{fig:view_allocation_qualitative}, all-view supervision propagates
blur into thin structures and object boundaries, while RVA yields sharper
contours and more stable local appearance. The initial-scene rows in
Table~\ref{tab:main_comparison} show the same trend: RVA-selected
$\mathcal{G}_0$ reduces LPIPS by 0.011, 0.013, and 0.015 and improves MUSIQ by
1.71, 2.49, and 4.11 from slight to severe blur. CLIP-IQA remains comparable,
with an equal score under slight blur and a small gain under severe blur. These
results show that RVA reduces blur-related degradation and provides a cleaner
initialization for repair.

\subsection{Quantitative Comparison}

Table~\ref{tab:main_comparison} reports perceptual results on GS2E and GSOTM, together with all-view and selected-view initial scenes to separate view allocation from repair. For a fair comparison, all methods use the same input sequences and are evaluated at the same observed input poses against the same sharp targets. Under this protocol, ClearGS improves over DiFiX3D on all GS2E blur levels, reducing LPIPS by 0.016, 0.030, and 0.038 and increasing CLIP-IQA by 0.030, 0.032, and 0.040 from slight to severe blur. MUSIQ gains also grow with blur strength, from 3.14 to 6.25. On GSOTM, ClearGS achieves the best CLIP-IQA and MUSIQ in all four settings, and its LPIPS is best in three settings and within 0.001 of the best score under rolling shutter.

\subsection{Qualitative Comparison}

\begin{figure}[!t]
\centering
\includegraphics[width=0.90\linewidth]{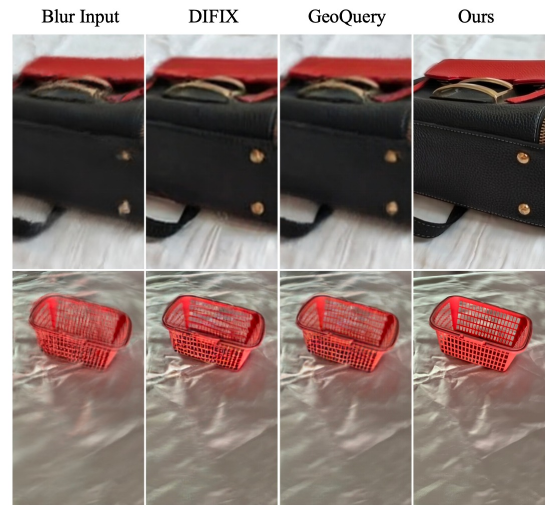}
\caption{Qualitative comparison of repair outputs on two GS2E scenes. Each row shows the same camera pose. From left to right, the panels show the common pre-repair 3DGS rendering, the repairs produced by DiFiX3D and GeoQuery, and the repair selected by ClearGS.}
\label{fig:repair_qualitative_comparison}
\end{figure}

\begin{table}[!t]
\centering
\caption{Ablation of RIVR candidate selection. ``w/o Obs.'' replaces
the raw video observation $O_t$ with the current render $R_t$ as the
input to Restormer, while ``w/o $F_t$'' removes the high-frequency
fused candidate. The remaining repair-distillation pipeline is unchanged.}
\label{tab:rivr_candidate_ablation}
\scriptsize
\setlength{\tabcolsep}{2.0pt}
\renewcommand{\arraystretch}{1.05}
\resizebox{\linewidth}{!}{%
\begin{tabular}{@{}llccccccc@{}}
\toprule
\multirow{2}{*}{Metric} & \multirow{2}{*}{Variant}
& \multicolumn{3}{c}{GS2E}
& \multicolumn{4}{c}{GSOTM} \\
\cmidrule(lr){3-5}
\cmidrule(l){6-9}
& & S & M & V & MB & RS & MB+RS & PN \\
\midrule
\multirow{3}{*}{LPIPS$\downarrow$}
& w/o Obs. & 0.385 & 0.401 & 0.414 & 0.247 & 0.294 & 0.358 & 0.398 \\
& w/o $F_t$ & 0.336 & 0.348 & 0.365 & \cellcolor{bestbg}\textbf{0.171} & \cellcolor{bestbg}\textbf{0.213} & \cellcolor{bestbg}\textbf{0.298} & \cellcolor{bestbg}\textbf{0.196} \\
& Full RIVR & \cellcolor{bestbg}\textbf{0.333} & \cellcolor{bestbg}\textbf{0.342} & \cellcolor{bestbg}\textbf{0.350} & 0.193 & 0.225 & 0.302 & 0.197 \\
\midrule
\multirow{3}{*}{CLIP-IQA$\uparrow$}
& w/o Obs. & 0.283 & 0.270 & 0.279 & 0.217 & 0.239 & 0.209 & 0.258 \\
& w/o $F_t$ & 0.280 & 0.275 & 0.282 & 0.263 & 0.280 & 0.221 & \cellcolor{bestbg}\textbf{0.269} \\
& Full RIVR & \cellcolor{bestbg}\textbf{0.288} & \cellcolor{bestbg}\textbf{0.287} & \cellcolor{bestbg}\textbf{0.293} & \cellcolor{bestbg}\textbf{0.314} & \cellcolor{bestbg}\textbf{0.296} & \cellcolor{bestbg}\textbf{0.243} & 0.267 \\
\midrule
\multirow{3}{*}{MUSIQ$\uparrow$}
& w/o Obs. & 37.15 & 35.59 & 34.98 & 47.37 & 57.30 & 46.20 & 47.13 \\
& w/o $F_t$ & 43.43 & 42.12 & 40.67 & 51.64 & 63.30 & 45.53 & 58.98 \\
& Full RIVR & \cellcolor{bestbg}\textbf{44.67} & \cellcolor{bestbg}\textbf{43.57} & \cellcolor{bestbg}\textbf{43.18} & \cellcolor{bestbg}\textbf{62.03} & \cellcolor{bestbg}\textbf{64.25} & \cellcolor{bestbg}\textbf{52.64} & \cellcolor{bestbg}\textbf{59.07} \\
\bottomrule
\end{tabular}%
}
\end{table}

\begin{figure*}[!t]
\centering
\includegraphics[width=0.93\textwidth]{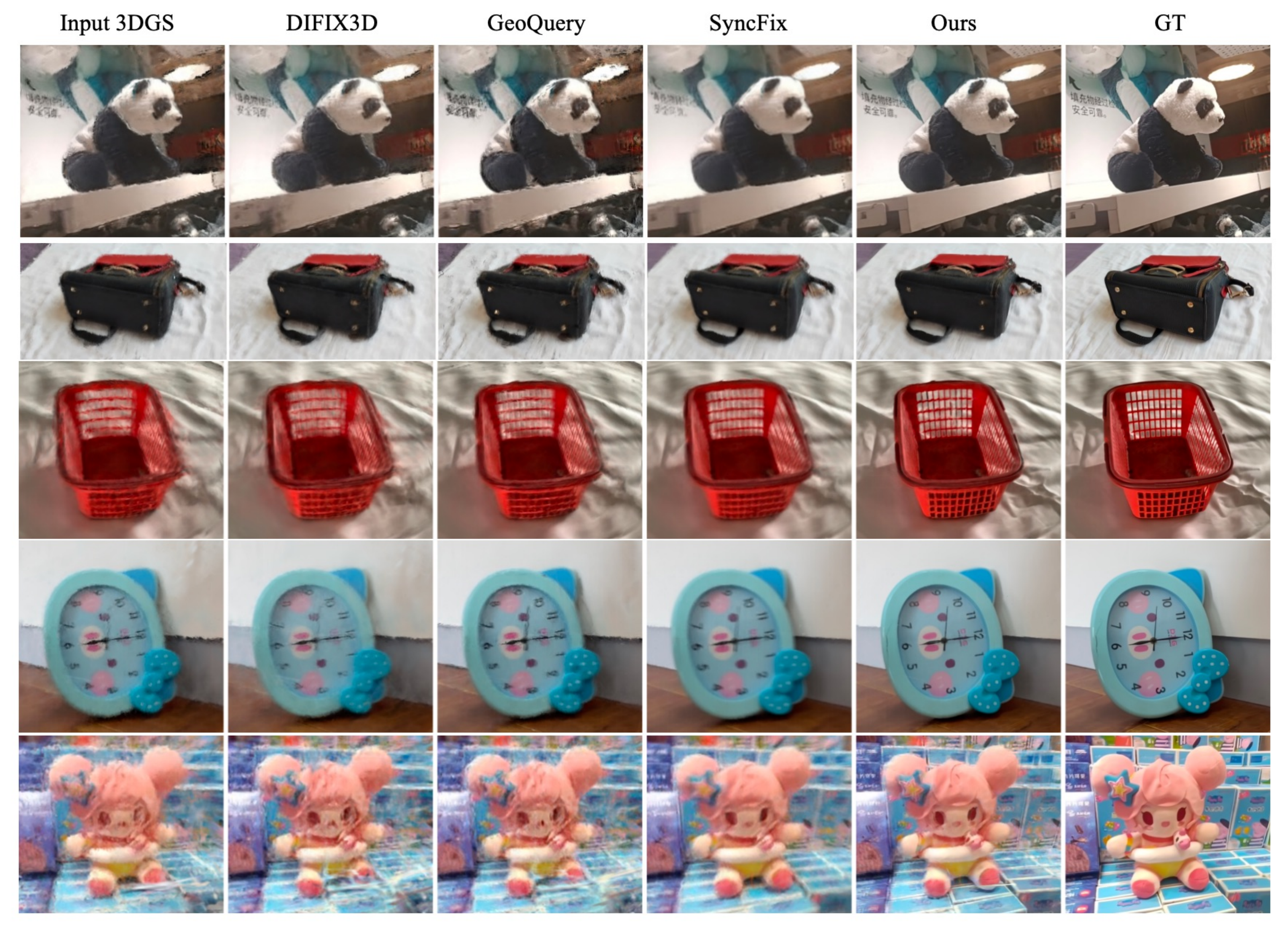}
\caption{Qualitative comparison on five GS2E scenes. From left to right: the input 3DGS, DiFiX3D, GeoQuery, SyncFix, ClearGS, and the ground truth. ClearGS recovers sharper object boundaries and local details while preserving the scene appearance.}
\label{fig:main_qualitative_comparison}
\end{figure*}

Figure~\ref{fig:repair_qualitative_comparison} compares repair outputs at the
same observed camera poses, while Figure~\ref{fig:main_qualitative_comparison}
compares the final renderings with the input 3DGS, existing restoration
baselines, and sharp ground truth. The pre-repair renderings retain visible
blur and weakened local structures, and the restoration baselines improve
sharpness but can still leave residual artifacts or less stable scene details.
In contrast, ClearGS produces cleaner object boundaries and more coherent
local textures in the shown examples. These visual differences align with the
quantitative results in Table~\ref{tab:main_comparison}, where ClearGS obtains
the best overall perceptual scores and larger gains under stronger
degradations.

\subsection{Ablation Study}

\subsubsection{Coverage Allocation in RVA}
Table~\ref{tab:rva_coverage_ablation} evaluates the coverage top-up step in RVA. On GS2E, full RVA improves PSNR by 0.538 and 0.512 and SSIM by 0.017 and 0.019 under medium and severe blur, where reliability weighting is more likely to suppress useful but degraded trajectory regions. It also lowers LPIPS under slight and medium blur, while matching the no-top-up variant under severe blur. Under slight blur, the small PSNR drop and nearly unchanged SSIM suggest limited gain when most views remain reliable. On GSOTM, the effect is mixed but improves several settings, including RS and PN. These results show that coverage top-up complements reliability weighting by recovering useful trajectory evidence under stronger degradation.

\subsubsection{RIVR Candidate Selection}
Table~\ref{tab:rivr_candidate_ablation} ablates the RIVR candidate pool. The full model selects among the current render $R_t$, the no-reference Restormer output from the aligned video observation $O_t$, and the high-frequency fusion candidate $F_t$. In w/o $O_t$, Restormer is still used, but its input is changed from $O_t$ to $R_t$; thus the ablation removes video evidence rather than the repair module. Removing this candidate consistently hurts LPIPS, CLIP-IQA, and MUSIQ on both datasets, with large MUSIQ drops on GSOTM motion blur and pose noise (14.66 and 11.94). Removing $F_t$ generally lowers MUSIQ and CLIP-IQA, indicating that the fused candidate contributes perceptually preferred high-frequency details.

\section{Conclusion and Discussion}

We presented ClearGS for 3DGS reconstruction from handheld videos with mixed frame quality and uneven trajectory coverage. ClearGS assigns graded supervision through RVA, separating visual reliability, degradation risk, and geometric utility while preserving coverage in weakly observed trajectory regions. It then performs no-reference in-video restoration: the current render serves as a pose-aligned structural anchor, and the corresponding raw video frame provides the evidence to be restored, without requiring matched clean references. A final consolidation pass revisits accepted repairs across the trajectory to preserve details learned early. Experiments on GS2E and GSOTM show improved perceptual quality, sharper boundaries, and more stable textures. Future work will study degradation-specific restoration and geometric checks for repaired evidence.

\bibliography{aaai2027}

@inproceedings{wu2025difix3d+,
  title={Difix3d+: Improving 3d reconstructions with single-step diffusion models},
  author={Wu, Jay Zhangjie and Zhang, Yuxuan and Turki, Haithem and Ren, Xuanchi and Gao, Jun and Shou, Mike Zheng and Fidler, Sanja and Gojcic, Zan and Ling, Huan},
  booktitle={Proceedings of the IEEE/CVF Conference on Computer Vision and Pattern Recognition},
  pages={26024--26035},
  year={2025}
}

@inproceedings{wu2025genfusion,
  title={Genfusion: Closing the loop between reconstruction and generation via videos},
  author={Wu, Sibo and Xu, Congrong and Huang, Binbin and Geiger, Andreas and Chen, Anpei},
  booktitle={Proceedings of the Computer Vision and Pattern Recognition Conference},
  pages={6078--6088},
  year={2025}
}

@article{yin2025gsfixer,
  title={Gsfixer: Improving 3d gaussian splatting with reference-guided video diffusion priors},
  author={Yin, Xingyilang and Zhang, Qi and Chang, Jiahao and Feng, Ying and Fan, Qingnan and Yang, Xi and Pun, Chi-Man and Zhang, Huaqi and Cun, Xiaodong},
  journal={arXiv preprint arXiv:2508.09667},
  year={2025}
}

@inproceedings{zhu2026gaussfusion,
  title={GaussFusion: Improving 3D Reconstruction in the Wild with A Geometry-Informed Video Generator},
  author={Zhu, Liyuan and Narayana, Manjunath and Stary, Michal and Hutchcroft, Will and Wetzstein, Gordon and Armeni, Iro},
  booktitle={Proceedings of the IEEE/CVF Conference on Computer Vision and Pattern Recognition},
  pages={15432--15442},
  year={2026}
}

@article{li2026syncfix,
  title={SyncFix: Fixing 3D Reconstructions via Multi-View Synchronization},
  author={Li, Deming and Yadav, Abhay and Peng, Cheng and Chellappa, Rama and Bhattad, Anand},
  journal={arXiv preprint arXiv:2604.11797},
  year={2026}
}

@article{cao2026geoquery,
  title={GeoQuery: Geometry-Query Diffusion for Sparse-View Reconstruction},
  author={Cao, Xiao and Li, Yuze and Zhang, Youmin and Song, Jiayu and Yan, Cheng and Li, Wen and Duan, Lixin},
  journal={arXiv preprint arXiv:2605.12399},
  year={2026}
}

@article{ni2025g4splat,
  title={G4Splat: Geometry-Guided Gaussian Splatting with Generative Prior},
  author={Ni, Junfeng and Chen, Yixin and Yang, Zhifei and Liu, Yu and Lu, Ruijie and Zhu, Song-Chun and Huang, Siyuan},
  journal={arXiv preprint arXiv:2510.12099},
  year={2025}
}

@article{liu20243dgs,
  title={3dgs-enhancer: Enhancing unbounded 3d gaussian splatting with view-consistent 2d diffusion priors},
  author={Liu, Xi and Zhou, Chaoyi and Huang, Siyu},
  journal={Advances in Neural Information Processing Systems},
  volume={37},
  pages={133305--133327},
  year={2024}
}

@inproceedings{wu2024reconfusion,
  title={Reconfusion: 3d reconstruction with diffusion priors},
  author={Wu, Rundi and Mildenhall, Ben and Henzler, Philipp and Park, Keunhong and Gao, Ruiqi and Watson, Daniel and Srinivasan, Pratul P and Verbin, Dor and Barron, Jonathan T and Poole, Ben and others},
  booktitle={Proceedings of the IEEE/CVF conference on computer vision and pattern recognition},
  pages={21551--21561},
  year={2024}
}

@article{kerbl20233d,
  title={3d gaussian splatting for real-time radiance field rendering.},
  author={Kerbl, Bernhard and Kopanas, Georgios and Leimk{\"u}hler, Thomas and Drettakis, George and others},
  journal={ACM Trans. Graph.},
  volume={42},
  number={4},
  pages={139--1},
  year={2023}
}

@inproceedings{chung2024depth,
  title={Depth-regularized optimization for 3d gaussian splatting in few-shot images},
  author={Chung, Jaeyoung and Oh, Jeongtaek and Lee, Kyoung Mu},
  booktitle={Proceedings of the IEEE/CVF Conference on Computer Vision and Pattern Recognition},
  pages={811--820},
  year={2024}
}

@article{xiong2023sparsegs,
  title={Sparsegs: Real-time 360 sparse view synthesis using gaussian splatting},
  author={Xiong, Haolin and Muttukuru, Sairisheek and Upadhyay, Rishi and Chari, Pradyumna and Kadambi, Achuta},
  journal={arXiv e-prints},
  pages={arXiv--2312},
  year={2023}
}

@inproceedings{zhu2024fsgs,
  title={Fsgs: Real-time few-shot view synthesis using gaussian splatting},
  author={Zhu, Zehao and Fan, Zhiwen and Jiang, Yifan and Wang, Zhangyang},
  booktitle={European conference on computer vision},
  pages={145--163},
  year={2024},
  organization={Springer}
}

@inproceedings{zhang2024cor,
  title={Cor-gs: sparse-view 3d gaussian splatting via co-regularization},
  author={Zhang, Jiawei and Li, Jiahe and Yu, Xiaohan and Huang, Lei and Gu, Lin and Zheng, Jin and Bai, Xiao},
  booktitle={European conference on computer vision},
  pages={335--352},
  year={2024},
  organization={Springer}
}

@inproceedings{park2025dropgaussian,
  title={Dropgaussian: Structural regularization for sparse-view gaussian splatting},
  author={Park, Hyunwoo and Ryu, Gun and Kim, Wonjun},
  booktitle={Proceedings of the computer vision and pattern recognition conference},
  pages={21600--21609},
  year={2025}
}

@article{chen2026quantifying,
  title={Quantifying and alleviating co-adaptation in sparse-view 3d gaussian splatting},
  author={Chen, Kangjie and Zhong, Yingji and Li, Zhihao and Lin, Jiaqi and Chen, Youyu and Qin, Minghan and Wang, Haoqian},
  journal={Advances in Neural Information Processing Systems},
  volume={38},
  pages={115939--115968},
  year={2026}
}

@article{song2025d,
  title={{D$^2$GS}: Depth-and-Density Guided Gaussian Splatting for Stable and Accurate Sparse-View Reconstruction},
  author={Song, Meixi and Lin, Xin and Zhang, Dizhe and Li, Haodong and Li, Xiangtai and Du, Bo and Qi, Lu},
  journal={arXiv preprint arXiv:2510.08566},
  year={2025}
}

@article{shen2026sparse,
  title={Sparse-to-Complete: From Sparse Image Captures to Complete 3D Scenes},
  author={Shen, Yiyang and Yang, Yin and Zhou, Kun and Shao, Tianjia},
  journal={arXiv preprint arXiv:2605.05664},
  year={2026}
}

@inproceedings{jiang2026freescale,
  title={FreeScale: Scaling 3D Scenes via Certainty-Aware Free-View Generation},
  author={Jiang, Chenhan and Chen, Yu and Zhang, Qingwen and Song, Jifei and Xu, Songcen and Yeung, Dit-Yan and Deng, Jiankang},
  booktitle={Proceedings of the IEEE/CVF Conference on Computer Vision and Pattern Recognition},
  pages={330--340},
  year={2026}
}

@inproceedings{li2024dngaussian,
  title={Dngaussian: Optimizing sparse-view 3d gaussian radiance fields with global-local depth normalization},
  author={Li, Jiahe and Zhang, Jiawei and Bai, Xiao and Zheng, Jin and Ning, Xin and Zhou, Jun and Gu, Lin},
  booktitle={Proceedings of the IEEE/CVF conference on computer vision and pattern recognition},
  pages={20775--20785},
  year={2024}
}

@inproceedings{xu2024mvpgs,
  title={Mvpgs: Excavating multi-view priors for gaussian splatting from sparse input views},
  author={Xu, Wangze and Gao, Huachen and Shen, Shihe and Peng, Rui and Jiao, Jianbo and Wang, Ronggang},
  booktitle={European Conference on Computer Vision},
  pages={203--220},
  year={2024},
  organization={Springer}
}

@inproceedings{ma2022deblur,
  title={Deblur-nerf: Neural radiance fields from blurry images},
  author={Ma, Li and Li, Xiaoyu and Liao, Jing and Zhang, Qi and Wang, Xuan and Wang, Jue and Sander, Pedro V},
  booktitle={Proceedings of the IEEE/CVF conference on computer vision and pattern recognition},
  pages={12861--12870},
  year={2022}
}

@inproceedings{lee2023dp,
  title={Dp-nerf: Deblurred neural radiance field with physical scene priors},
  author={Lee, Dogyoon and Lee, Minhyeok and Shin, Chajin and Lee, Sangyoun},
  booktitle={Proceedings of the IEEE/CVF Conference on Computer Vision and Pattern Recognition},
  pages={12386--12396},
  year={2023}
}

@inproceedings{wang2023bad,
  title={Bad-nerf: Bundle adjusted deblur neural radiance fields},
  author={Wang, Peng and Zhao, Lingzhe and Ma, Ruijie and Liu, Peidong},
  booktitle={Proceedings of the IEEE/CVF Conference on Computer Vision and Pattern Recognition},
  pages={4170--4179},
  year={2023}
}

@inproceedings{lee2023exblurf,
  title={Exblurf: Efficient radiance fields for extreme motion blurred images},
  author={Lee, Dongwoo and Oh, Jeongtaek and Rim, Jaesung and Cho, Sunghyun and Lee, Kyoung Mu},
  booktitle={Proceedings of the IEEE/CVF International Conference on Computer Vision},
  pages={17639--17648},
  year={2023}
}

@inproceedings{lee2024deblurring,
  title={Deblurring 3d gaussian splatting},
  author={Lee, Byeonghyeon and Lee, Howoong and Sun, Xiangyu and Ali, Usman and Park, Eunbyung},
  booktitle={European Conference on Computer Vision},
  pages={127--143},
  year={2024},
  organization={Springer}
}

@inproceedings{zhao2024bad,
  title={Bad-gaussians: Bundle adjusted deblur gaussian splatting},
  author={Zhao, Lingzhe and Wang, Peng and Liu, Peidong},
  booktitle={European Conference on Computer Vision},
  pages={233--250},
  year={2024},
  organization={Springer}
}

@inproceedings{peng2024bags,
  title={Bags: Blur agnostic gaussian splatting through multi-scale kernel modeling},
  author={Peng, Cheng and Tang, Yutao and Zhou, Yifan and Wang, Nengyu and Liu, Xijun and Li, Deming and Chellappa, Rama},
  booktitle={European Conference on Computer Vision},
  pages={293--310},
  year={2024},
  organization={Springer}
}

@inproceedings{lee2025comogaussian,
  title={Comogaussian: Continuous motion-aware gaussian splatting from motion-blurred images},
  author={Lee, Jungho and Kim, Donghyeong and Lee, Dogyoon and Cho, Suhwan and Lee, Minhyeok and Lee, Wonjoon and Kim, Taeoh and Wee, Dongyoon and Lee, Sangyoun},
  booktitle={Proceedings of the IEEE/CVF International Conference on Computer Vision},
  pages={26415--26424},
  year={2025}
}

@inproceedings{zhang2026unblur,
  title={Unblur-SLAM: Dense Neural SLAM for Blurry Inputs},
  author={Zhang, Qi and Rozumny, Denis and Girlanda, Francesco and Karaoglu, Sezer and Pollefeys, Marc and Gevers, Theo and Oswald, Martin R},
  booktitle={Proceedings of the IEEE/CVF Conference on Computer Vision and Pattern Recognition},
  pages={352--362},
  year={2026}
}

@inproceedings{ke2021musiq,
  title={Musiq: Multi-scale image quality transformer},
  author={Ke, Junjie and Wang, Qifei and Wang, Yilin and Milanfar, Peyman and Yang, Feng},
  booktitle={Proceedings of the IEEE/CVF international conference on computer vision},
  pages={5148--5157},
  year={2021}
}

@inproceedings{li2025gs2e,
  title={GS2E: Gaussian Splatting is an Effective Data Generator for Event Stream Generation},
  author={Li, Yuchen and Feng, Chaoran and Tang, Zhenyu and Deng, Kaiyuan and Yu, Wangbo and Tian, Yonghong and Yuan, Li},
  booktitle={Advances in Neural Information Processing Systems},
  volume={38},
  year={2025}
}

@inproceedings{seiskari2024gaussian,
  title={Gaussian splatting on the move: Blur and rolling shutter compensation for natural camera motion},
  author={Seiskari, Otto and Ylilammi, Jerry and Kaatrasalo, Valtteri and Rantalankila, Pekka and Turkulainen, Matias and Kannala, Juho and Rahtu, Esa and Solin, Arno},
  booktitle={European conference on computer vision},
  pages={160--177},
  year={2024},
  organization={Springer}
}

@inproceedings{xue2026uncertainty,
  title={Uncertainty-driven 3D Gaussian Splatting Active Mapping via Anisotropic Visibility Field},
  author={Xue, Shangjie and Dill, Jesse and Ahuja, Dhruv and Dellaert, Frank and Tsiotras, Panagiotis and Xu, Danfei},
  booktitle={Proceedings of the IEEE/CVF Conference on Computer Vision and Pattern Recognition},
  pages={5014--5026},
  year={2026}
}

@inproceedings{chen2026coverage,
  title={Coverage Optimization for Camera View Selection},
  author={Chen, Timothy and Dai, Adam and Adang, Maximilian and Gao, Grace and Schwager, Mac},
  booktitle={Proceedings of the IEEE/CVF Conference on Computer Vision and Pattern Recognition},
  pages={19443--19451},
  year={2026}
}

@article{polyzos2025activeinitsplat,
  title={Activeinitsplat: How active image selection helps gaussian splatting},
  author={Polyzos, Konstantinos D and Bacharis, Athanasios and Madhuvarasu, Saketh and Papanikolopoulos, Nikos and Javidi, Tara},
  journal={arXiv preprint arXiv:2503.06859},
  year={2025}
}

@article{bai2025qwen3,
  title={Qwen3-vl technical report},
  author={Bai, Shuai and Cai, Yuxuan and Chen, Ruizhe and Chen, Keqin and Chen, Xionghui and Cheng, Zesen and Deng, Lianghao and Ding, Wei and Gao, Chang and Ge, Chunjiang and others},
  journal={arXiv preprint arXiv:2511.21631},
  year={2025}
}

@inproceedings{zamir2022restormer,
  title={Restormer: Efficient transformer for high-resolution image restoration},
  author={Zamir, Syed Waqas and Arora, Aditya and Khan, Salman and Hayat, Munawar and Khan, Fahad Shahbaz and Yang, Ming-Hsuan},
  booktitle={Proceedings of the IEEE/CVF conference on computer vision and pattern recognition},
  pages={5728--5739},
  year={2022}
}


\end{document}